\documentclass[11pt,a4paper]{article}
\usepackage[hyperref]{acl2021}
\usepackage{times}
\usepackage{latexsym}

\usepackage{graphicx}
\usepackage{amssymb}
\usepackage{amsmath}
\usepackage{booktabs}
\usepackage{multirow}

\usepackage{enumitem}
\setenumerate[1]{itemsep=0pt,partopsep=0pt,parsep=\parskip,topsep=5pt}
\setitemize[1]{itemsep=0pt,partopsep=0pt,parsep=\parskip,topsep=5pt}
\setdescription{itemsep=0pt,partopsep=0pt,parsep=\parskip,topsep=5pt}

\usepackage{microtype}

\aclfinalcopy 

\title{Link Prediction on N-ary Relational Facts: A Graph-based Approach}
  
\author{{\bf Quan Wang, Haifeng Wang, Yajuan Lyu, Yong Zhu} \\
  Baidu Inc., Beijing, China \\
  {\tt \{wangquan05,wanghaifeng,lvyajuan,zhuyong\}@baidu.com}}

\date{}

\begin{document}
\maketitle
\begin{abstract}
Link prediction on knowledge graphs (KGs) is a key research topic. Previous work mainly focused on binary relations, paying less attention to higher-arity relations although they are ubiquitous in real-world KGs. This paper considers link prediction upon n-ary relational facts and proposes a graph-based approach to this task. The key to our approach is to represent the n-ary structure of a fact as a small heterogeneous graph, and model this graph with edge-biased fully-connected attention. $\!$The fully-connected attention captures universal inter-vertex interactions, while with edge-aware attentive biases to particularly encode the graph structure and its heterogeneity. In this fashion, our approach fully models global and local dependencies in each n-ary fact, and hence can more effectively capture associations therein. Extensive evaluation verifies the effectiveness and superiority of our approach. It performs substantially and consistently better than current state-of-the-art across a variety of n-ary relational benchmarks. Our code is publicly available.\footnote{\url{https://github.com/PaddlePaddle/Research/tree/master/KG/ACL2021\_GRAN}}
\end{abstract}

\section{Introduction}
Web-scale knowledge graphs (KGs), such as Freebase~\cite{bollacker2008:Freebase}, Wikidata \cite{vrandevcic2014:Wikidata}, and Google Knowledge Vault \cite{dong2014:KnowledgeVault},  are useful resources for many real-world applications, ranging from Web search and question answering to recommender systems. Though impressively large, these modern KGs are still known to be greatly incomplete and missing crucial facts \cite{west2014:Incomplete}. Link prediction which predicts missing links in KGs has therefore become an important research topic.

Previous studies mainly consider link prediction upon binary relational facts, which encode binary relations between pairs of entities and are usually represented as ($subject$, $relation$, $object$) triples. Nevertheless, besides binary relational facts, n-ary relational facts that involve more than two entities are also ubiquitous in reality, e.g., {\it Marie Curie received Nobel Prize in Physics in 1903 together with Pierre Curie and Antoine Henri Becquerel} is a typical 5-ary fact. As pointed out by \citeauthor{wen2016:m-transh} \shortcite{wen2016:m-transh}, more than $1/3$ of the entities in Freebase actually participate in n-ary relational facts. 

\begin{figure}[t]
\centering
\includegraphics[width=0.48 \textwidth]{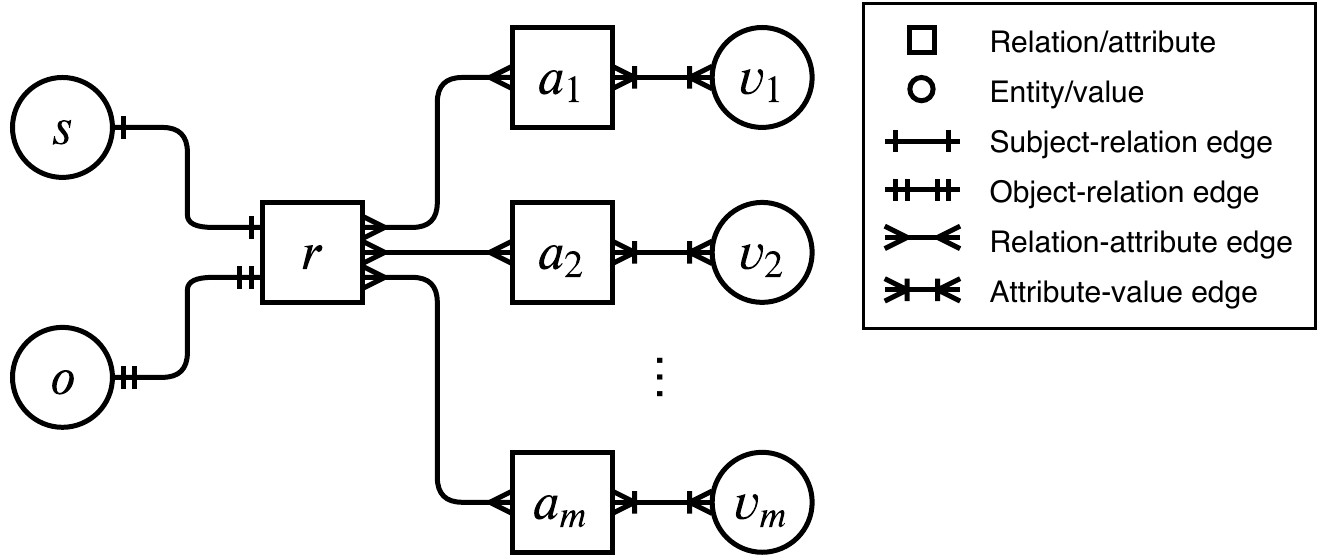}\\
\caption{An n-ary fact as a heterogenous graph, with relations/attributes and entities/values as vertices, and four types of edges designed between the vertices.}
\label{fig:hete-graph}
\end{figure} 

Despite the ubiquitousness, only a few studies have examined link prediction on n-ary relational facts. In these studies, an n-ary fact is typically represented as a set of peer attributes (relations) along with their values (entities), e.g., \{{\it person: Marie Curie}, {\it award: Nobel Prize in Physics}, {\it point-in-time: 1903}, {\it together-with: Pierre Curie}, {\it together-with: Antoine Henri Becquerel}\}. Link prediction then is achieved by learning the relatedness either between the values \cite{zhang2018:rae,liu2020:getd,fatemi2020:hype} or between the attribute-value pairs \cite{guan2019:nalp,liu2021:ram}. $\!$This representation inherently assumes that attributes of a same n-ary fact are equally important, which is usually not the case. To further discriminate importance of different attributes, \citeauthor{rosso2020:hinge} \shortcite{rosso2020:hinge} and \citeauthor{guan2020:NeuInfer} \shortcite{guan2020:NeuInfer} later proposed to represent an n-ary fact as a primary triple coupled with auxiliary attribute-value descriptions, e.g., in the above 5-ary fact, ({\it Marie Curie}, {\it award-received}, {\it Nobel Prize in Physics}) is the primary triple and {\it point-in-time: 1903}, {\it together-with: Pierre Curie}, {\it together-with: Antoine Henri Becquerel} are auxiliary descriptions. Link prediction then is achieved by measuring the validity of the primary triple and its compatibility with each attribute-value pair. These attribute-value pairs, however, are modeled independently before a final aggregation, thus ignoring intrinsic semantic relatedness in between.

This work in general follows \citeauthor{rosso2020:hinge} \shortcite{rosso2020:hinge} and \citeauthor{guan2020:NeuInfer} \shortcite{guan2020:NeuInfer}'s expressive representation form of n-ary facts, but takes a novel graph learning perspective for modeling and reasoning with such facts. Given an n-ary fact represented as a primary subject-relation-object triple $(s, r, o)$ with auxiliary attribute-value pairs $\{(a_i \negmedspace:\negmedspace v_i)\}$, we first formalize the fact as a heterogenous graph. This graph, as we illustrate in Figure~\ref{fig:hete-graph}, takes relations and entities (attributes and values) as vertices, and introduces four types of edges, i.e., subject-relation, object-relation, relation-attribute, and attribute-value, to denote distinct connectivity patterns between these vertices. In this fashion, the full semantics of the given fact will be retained in the graph. Then, based on this graph representation, we employ a fully-connected attention module to characterize inter-vertex interactions, while further introducing edge-aware attentive biases to particularly handle the graph structure and heterogeneity. This enables us to capture not only local but also global dependencies within the fact. Our approach directly encodes each n-ary fact as a whole graph so as to better capture rich associations therein. In this sense, we call it GRAph-based N-ary relational learning (GRAN). 

The most similar prior art to this work is STARE \cite{galkin2020:stare}, which uses a message passing based graph encoder to obtain relation (attribute) and entity (value) embeddings, and feeds these embeddings into a Transformer \cite{vaswani2017:transformer} decoder to score n-ary facts. Our approach is more neatly designed by (1) excluding the computational-heavy graph encoder which, according to a contemporaneous study \cite{yu2021:hy-transformer}, may not be necessary given an expressive enough decoder, and (2) modeling the full n-ary structure of a fact during decoding which enables to capture not only global but also local dependencies therein. 

We evaluate our approach on a variety of n-ary link prediction benchmarks. Experimental results reveal that GRAN works particularly well in learning and reasoning with n-ary relational facts, consistently and substantially outperforming current state-of-the-art across all the benchmarks. $\!$Our main contributions are summarized as follows:
\begin{itemize}
\item We present a novel graph-based approach to learning and reasoning with n-ary facts, capable of capturing rich associations therein.
\item We demonstrate the effectiveness and superiority of our approach, establishing new state-of-the-art across a variety of benchmarks.
\end{itemize} 

\section{Problem statement}
This section formally defines n-ary relational facts and the link prediction task on this kind of data.
\paragraph{Definition 1 (N-ary relational fact)}
An n-ary relational fact $\mathcal{F}$ is a primary subject-relation-object triple $(s,r,o)$ coupled with $m$ auxiliary attribute-value pairs $\{(a_i \negmedspace:\negmedspace v_i)\}_{i=1}^m$, where $r, a_1, \cdots, a_m \in \mathcal{R}$ and $s, o, v_1, \cdots, v_m \in \mathcal{E}$, with $\mathcal{R}$ and $\mathcal{E}$ being the sets of relations and entities, respectively. We slightly abuse terminology here by referring to the primary relation and all attributes as relations, and referring to the subject, object, and values as entities unless otherwise specified. The arity of the fact is $(m+$ $2)$, i.e., the number of entities in the fact. 

\paragraph{Definition 2 (N-ary link prediction)}
N-ary link prediction aims to predict a missing element from an n-ary fact. The missing element can be either an entity $\in \{s, o, v_1, \cdots, v_m\}$ or a relation $\in \{r, a_1,$ $\cdots, a_m\}$, e.g., to predict the primary subject of the incomplete n-ary fact $\big((?, r, o), \{(a_i \negmedspace:\negmedspace v_i)\}_{i=1}^m\big)$. 

\section{Graph-based n-ary relational learning}
This section presents GRAN, our graph-based approach to n-ary link prediction. There are two key factors of our approach: graph representation and graph learning. The former represents n-ary facts as graphs, and the latter learns with these graphs to perform inference on n-ary facts.

\subsection{Graph representation}
We elaborate the first key factor: graph representation of n-ary facts. Given an n-ary fact defined as $\mathcal{F}=\big((s, r, o), \{(a_i \negmedspace:\negmedspace v_i)\}_{i=1}^m\big)$, we reformulate it equivalently as a heterogeneous graph $\mathcal{G}=(\mathcal{V}, \mathcal{L})$. The vertex set $\mathcal{V}$ consists of all entities and relations in the fact, i.e., $\mathcal{V}=\{r, s, o, a_1, \cdots\!, a_m, v_1, \cdots\!,$ $v_m\}$. The link set $\mathcal{L}$ consists of $(2m+2)$ undirected \mbox{edges of four types between the vertices, i.e.,}
\begin{itemize}
\item 1 subject-relation edge $(s, r)$,
\item 1 object-relation edge $(o, r)$,
\item $m$ relation-attribute edges $\{(r, a_i)\}_{i=1}^m$,
\item $m$ attribute-value edges $\{(a_i, v_i)\}_{i=1}^m$.
\end{itemize}
The graph heterogeneity is reflected in that the vertices and links are both typed, with type mapping functions $\phi\!: \mathcal{V} \!\rightarrow\! \{\textit{entity}, \textit{relation}\}$ and $\psi\!: \mathcal{L} \!\rightarrow\!  \{ $ $\textit{subject-relation}, \textit{object-relation}, \textit{relation-attribute},$ $\textit{attribute-value}\}$, respectively. Figure~\ref{fig:hete-graph} provides a visual illustration of this heterogenous graph. 

As we can see, the graph representation retains the full semantics of a given fact. It also enables us to model the fact as a whole and capture all possible interactions therein, which, as we will show later in our experiments, is crucial for learning with n-ary relational facts.

\subsection{Graph learning}
\begin{figure}[t]
\centering
\includegraphics[width=0.49 \textwidth]{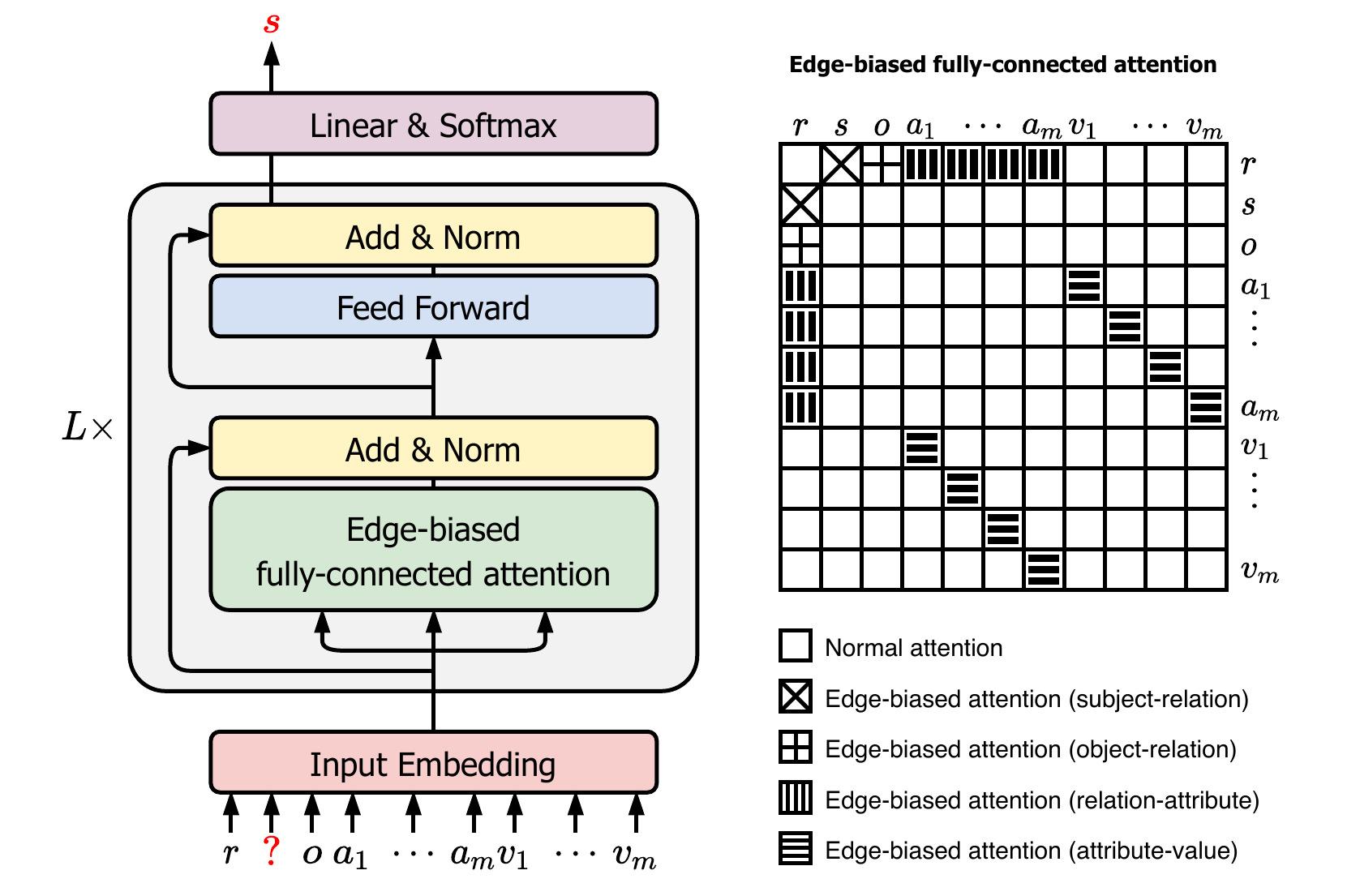}\\
\caption{Overview of the graph learning process, with edge-biased fully-connected attention illustrated.}
\label{fig:hete-gt}
\end{figure}

The second key factor is learning with heterogeneous graphs to perform inference on n-ary facts. Given an incomplete n-ary fact with a missing element, say $\big((?, r, o), \{(a_i \negmedspace:\negmedspace v_i)\}_{i=1}^m\big)$, which is represented as a heterogeneous graph, we feed the graph into an embedding layer, a stack of $L$ successive graph attention layers, and a final prediction layer to predict the missing element, say $s$. This whole process is sketched in Figure~\ref{fig:hete-gt} (left). 

The input embedding layer maps the elements of the input n-ary fact or, equivalently, the vertices of the input graph, to their continuous vector represen-tations (the missing element is denoted by a special token [MASK]). The $L$ graph attention layers then repeatedly encode the graph and update its vertex representations. Our graph attention generally inherits from Transformer \cite{vaswani2017:transformer} and its fully-connected attention which captures universal inter-vertex associations, but further introduces edge-aware attentive biases to particularly handle graph structure and heterogeneity. As such, we call it edge-biased fully-connected attention. After the graph encoding process, we use the representation of the special token [MASK] to predict the missing element. In the rest of this section, we emphasize the edge-biased fully-connected attention, and refer readers to \cite{vaswani2017:transformer} and Appendix~\ref{sec:append-gat} for other modules of our graph attention layer.

\paragraph{Edge-biased fully-connected attention}
We are given an input graph $\mathcal{G}=(\mathcal{V}, \mathcal{L})$, with vertex type mapping function $\phi$ and link type mapping function $\psi$. Vertices are associated with hidden states $(\mathbf{x}_1,$ $\cdots\!, \mathbf{x}_{|\mathcal{V}|}) \in \mathbb{R}^{d_x}$ generated by previous layers. The aim of this attention is to aggregate information from different vertices and update vertex representations, by taking into account the graph structure and its heterogeneity. We employ multi-head attention with $H$ heads, each applied independently to the input $(\mathbf{x}_1, \cdots\!, \mathbf{x}_{|\mathcal{V}|}) \in \mathbb{R}^{d_x}$ to generate updated vertex representations $(\mathbf{z}_1^h, \cdots\!, \mathbf{z}_{|\mathcal{V}|}^h) \in \mathbb{R}^{d_z}$ for $h$ $=1, \cdots\!, H$. These updated vertex representations are concatenated and linearly transformed to generate final attention output. We set $d_x=d$ and $d_z=$ $\frac{d}{H}$ for all layers and heads. Below we describe the specific design of each head, and we drop the head index $h$ for notational brevity.

Our attention follows the traditional query-key-value attention \cite{vaswani2017:transformer}. Specifically, for each input $\mathbf{x}_i$, we project it into a triple of query, key, and value as $(\mathbf{W}^Q \mathbf{x}_i, \mathbf{W}^K \mathbf{x}_i, \mathbf{W}^V \mathbf{x}_i) \in \mathbb{R}^{d_z}$, using parameters $\mathbf{W}^Q, \mathbf{W}^K, \mathbf{W}^V \in \mathbb{R}^{d_z \times d_x}$, respectively. Then we measure the similarity between each pair of vertices, say $i$ and $j$, as a scaled dot product of $i$'s query and $j$'s edge-biased key:
\begin{equation}\label{eq:attn-weights}
\small
\alpha_{ij} = \frac{(\mathbf{W}^Q \mathbf{x}_i)^\top (\mathbf{W}^K \mathbf{x}_j + \mathbf{e}_{ij}^K)}{\sqrt{d_z}}.
\end{equation}
After we obtain the similarity scores $\alpha_{ij}$, a softmax operation is applied, and the edge-biased values are aggregated accordingly to generate the updated representation for each vertex $i$:
\begin{equation}\label{eq:attn-output}
\small
\mathbf{z}_i = \sum_{j=1}^{|\mathcal{V}|} \frac{\exp{(\alpha_{ij})}}{\sum_{k=1}^{|\mathcal{V}|} \exp{(\alpha_{ik})}} (\mathbf{W}^V \mathbf{x}_j + \mathbf{e}_{ij}^V).
\end{equation}
We call this attention fully-connected as it takes into account similarity between any two vertices $i$ and $j$. We call it edge-biased as it further introduces attentive biases $\mathbf{e}_{ij}^K, \mathbf{e}_{ij}^V \in \mathbb{R}^{d_z}$ to encode the typed edge between $i$ and $j$, one to generate edge-biased key (cf. Eq.~(\ref{eq:attn-weights})) and the other edge-biased value (cf. Eq.~(\ref{eq:attn-output})). Introducing $\mathbf{e}_{ij}^K$ enables our attention to encode not only global dependencies that universally exist between any pair of vertices, but also local dependencies that are particularly indicated by typed edges. Introducing $\mathbf{e}_{ij}^V$ further propagates edge information to the attention output. If there is no edge linking $i$ and $j$  we set $\mathbf{e}_{ij}^K\!=\!\mathbf{e}_{ij}^V\!=\!\mathbf{0}$, which, at this time, degenerates to the conventional fully-connected attention used in Transformer \cite{vaswani2017:transformer}. As the attentive biases $\mathbf{e}_{ij}^K, \mathbf{e}_{ij}^V$ can be designed freely to meet any desired specifications, this attention is in essence quite flexible, capable of modeling arbitrary relationships between the input elements. This idea has actually been applied, e.g., to model relative positions between words within sentences \cite{shaw2018:relativeposition,wang2019:relativeposition}, or to model various kinds of mention dependencies for relation extraction \cite{xu2021:ssan}.

\paragraph{Edge-aware attentive biases}
We now elaborate how $\mathbf{e}_{ij}^K$ and $\mathbf{e}_{ij}^V$ are specifically designed for n-ary facts. Recall that given an n-ary fact represented as a heterogeneous graph $\mathcal{G}=(\mathcal{V}, \mathcal{L})$, there are 4 distinct types of edges in the graph: \textit{subject-relation}, \textit{object-relation}, \textit{relation-attribute}, \textit{attribute-value}. To each we assign a pair of key and value biases. The attentive biases between vertices $i$ and $j$ are then defined as the biases associated with the type of the edge linking $i$ and $j$:
\begin{equation}\label{eq:edges}
\small
(\mathbf{e}_{ij}^K,\!\mathbf{e}_{ij}^V) \!=\!
\begin{cases}
(\mathbf{0},\mathbf{0}),                 & \!\negmedspace\textrm{if } (i,\!j) \!\notin\! \mathcal{L}, \\
(\mathbf{e}_1^K,\!\mathbf{e}_1^V), & \!\negmedspace\textrm{if } \psi(i,\!j)\!=\!\textit{subject-relation}, \\
(\mathbf{e}_2^K,\!\mathbf{e}_2^V), & \!\negmedspace\textrm{if } \psi(i,\!j)\!=\!\textit{object-relation}, \\
(\mathbf{e}_3^K,\!\mathbf{e}_3^V), & \!\negmedspace\textrm{if } \psi(i,\!j)\!=\!\textit{relation-attribute}, \\
(\mathbf{e}_4^K,\!\mathbf{e}_4^V), & \!\negmedspace\textrm{if } \psi(i,\!j)\!=\!\textit{attribute-value}. \\
\end{cases}
\end{equation}
Here $\mathbf{e}_k^K, \mathbf{e}_k^V\in\mathbb{R}^{d_z}$ for $k=1,2,3,4$ are the key and value biases corresponding to the 4 edge types, shared across all layers and heads. In this way, the graph structure (whether there is an edge between two vertices) and its heterogeneity (which type the edge is between two vertices) can be well encoded into the attentive biases, and then propagated to the final attention output. Figure~\ref{fig:hete-gt} (right) visualizes the edge-biased attention between pairs of vertices in an n-ary fact. 

\subsection{Model training}
We directly use n-ary link prediction as our training task. Specifically, given an n-ary fact $\mathcal{F}\!\!=\!\! \big((s,r,o),$ $\{(a_i \negmedspace:\negmedspace v_i)\}_{i=1}^m \big)$ in the training set we create $(2m+3)$ training instances for it, each to predict a missing element (either an entity or a relation) given other elements in the fact, e.g., $\big((?, r, o), \{(a_i \negthickspace:\negthickspace v_i)\}_{i=1}^m\big)$ is to predict the primary subject and the answer to which is $s$. Here and in what follows we denote a training instance as $\widetilde{\mathcal{F}}$, with the missing element indicated by a special token [MASK]. This training instance is reformulated as a heterogeneous graph $\widetilde{\mathcal{G}}$ with vertices $(x_1, \cdots\!, x_k)$, where $k=2m+3$ is the total number of vertices therein. The label is denoted as $y$. We have $y\in\mathcal{E}$ for entity prediction and $y\in\mathcal{R}$ for relation prediction.

Each training instance $\widetilde{\mathcal{F}}$ or, equivalently, the corresponding graph $\widetilde{\mathcal{G}}$ is fed into the embedding, graph attention, and final prediction layers to predict the missing element, as we introduced above. Suppose after the successive graph attention layers we obtain  for the vertices $(x_1, \cdots\!, x_k)$ their hidden states $(\mathbf{h}_1, \cdots\!, \mathbf{h}_k) \in \mathbb{R}^d$. The hidden state corresponding to [MASK], denoted as $\mathbf{h}$ for brevity, is used for the final prediction. The prediction layer is constructed by two linear transformations followed by a standard softmax operation:
\begin{equation}\label{eq:prediction}
\small
\mathbf{p} = \textsc{Softmax} \big(\mathbf{W}_2^\top (\mathbf{W}_1 \mathbf{h} + \mathbf{b}_1) + \mathbf{b}_2\big).
\end{equation}
Here, we share $\mathbf{W}_2$ with the weight matrix of the input embedding layer, and $\mathbf{W}_1, \mathbf{b}_1, \mathbf{b}_2$ are freely learnable. The final output $\mathbf{p}$ is a probability distribution over entities in $\mathcal{E}$ or relations in $\mathcal{R}$, depending on the type of the missing element. 

We use the cross-entropy between the prediction and the label as our training loss:
\begin{equation}\label{eq:loss}
\small
L = \sum\nolimits_{t} y_t \log p_t,
\end{equation}
where $p_t$ is the $t$-th entry of the prediction $\mathbf{p}$, and $y_t$ the $t$-th entry of the label $\mathbf{y}$. As a one-hot label restricts each prediction task to a single answer, which might not be the case in practice, we employ label smoothing to lessen this restriction. Specifically, for entity prediction, we set $y_t\!=\!1\!-\!\epsilon^{(e)}$ for the target entity and $y_t\!=\!\frac{\epsilon^{(e)}}{|\mathcal{E}| -1}$ for each of the other entities, where $\epsilon^{(e)}$ is a small entity label smoothing rate. For relation prediction $y_t$ is set in a similar way, with relation label smoothing rate $\epsilon^{(r)}$. The loss is minimized using Adam optimizer \cite{kingma2014:adam}. We use learning rate warmup over the first 10\% training steps and linear decay of the learning rate. We also use batch normalization and dropout after each layer and sub-layer to regularize, stabilize, and speed up training.

Unlike previous methods which score individual facts and learn from positive-negative pairs \cite{rosso2020:hinge,guan2020:NeuInfer}, our training scheme bears two advantages: (1) Directly using n-ary link prediction as the training task can effectively avoid training-test discrepancy. (2) Introducing a special token [MASK] enables us to score a target element against all candidates simultaneously, which accelerates convergence during training and speeds up evaluation drastically \cite{dettmers2018:conve}.

\section{Experiments and results}
We evaluate GRAN in the link prediction task on n-ary facts. This section presents our experiments and results.

\begin{table*}[t]
\small\centering\setlength{\tabcolsep}{1pt}
\begin{tabular*}{1 \textwidth}{@{\extracolsep{\fill}}@{}lrrrrrrrc@{}}
\toprule
& All facts & Higher-arity facts (\%) & Entities & Relations & Train & Dev & Test & Arity \\
\midrule
JF17K               & 100,947 & 46,320 (45.9\%) & 28,645 & 501 & 76,379   & --          & 24,568 & 2-6 \\
WikiPeople        & 382,229 & 44,315 (11.6\%) & 47,765 & 193 & 305,725 & 38,223 & 38,281 & 2-9 \\
WikiPeople$^-$ & 369,866 & 9,482~~ (2.6\%) & 34,825 & 178 & 294,439 & 37,715 & 37,712 & 2-7 \\
\midrule
JF17K-3        & 34,544 & 34,544 (100\%) & 11,541 & 208 & 27,635   & 3,454   & 3,455   & 3 \\
JF17K-4        & 9,509   & 9,509 (100\%)   & 6,536   & 69   & 7,607     & 951      & 951      & 4 \\
WikiPeople-3 & 25,820 & 25,820 (100\%) & 12,270 & 112 & 20,656   & 2,582   & 2,582   & 3 \\
WikiPeople-4 & 15,188 & 15,188 (100\%) & 9,528   & 95   & 12,150   & 1,519   & 1,519   & 4 \\
\bottomrule
\end{tabular*}
\caption{Dataset statistics, where the columns respectively indicate the number of all facts, n-ary facts with $n>2$, entities, relations, facts in train/dev/test sets, and all possible arities.}\label{tab:datasets}
\end{table*}

\subsection{Datasets}
We consider standard n-ary link prediction benchmarks including:

{\bf JF17K}~\cite{zhang2018:rae}\footnote{\url{https://github.com/lijp12/SIR/}} is collected from Freebase. On this dataset, an n-ary relation is predefined by a set of attributes, and facts of this relation should have all corresponding values completely given. Take {\it music.group\_membership} as an example. All facts of this relation should get three values w.r.t. the predefined attributes, e.g., ({\it Guitar}, {\it Dean Fertita}, {\it Queens of the Stone Age}). The maximum arity of the relations there is $6$.

{\bf WikiPeople}~\cite{guan2019:nalp}\footnote{\url{https://github.com/gsp2014/NaLP}} is derived from Wikidata concerning entities of type {\it human}. On this dataset, n-ary facts are already represented as primary triples with auxiliary attribute-value pairs, which is more tolerant to data incompleteness. The maximum arity there is $9$. As the original dataset also contains literals, we follow \cite{rosso2020:hinge,galkin2020:stare} and consider another version that filters out statements containing literals. This filtered version is referred to as {\bf WikiPeople$^-$}, and the maximum arity there is $7$.

{\bf JF17K-3}, {\bf JF17K-4}, {\bf WikiPeople-3}, and {\bf Wiki-} {\bf People-4} \cite{liu2020:getd}\footnote{\url{https://github.com/liuyuaa/GETD}} are subsets of JF17K and WikiPeople, consisting solely of 3-ary and 4-ary relational facts therein, respectively. 

For JF17K and its subsets, we transform the representation of an n-ary fact to a primary triple coupled with auxiliary attribute-value pairs. We follow \cite{rosso2020:hinge,galkin2020:stare} and directly take the values corresponding to the first and second attributes as the primary subject and object, respectively. Other attributes and values are taken as auxiliary descriptions. Facts on each dataset are split into train/dev/test sets, and we use the original split. On JF17K which provides no dev set, we split 20\% of the train set for development. The statistics of these datasets are summarized in Table~\ref{tab:datasets}.

\subsection{Baseline methods}
We compare against the following state-of-the-art n-ary link prediction techniques:

{\bf RAE} \cite{zhang2018:rae} represents an n-ary fact as an $(n+1)$-tuple consisting of the predefined relation and its $n$ values. It generalizes a binary link prediction method TransH \cite{wang2014:transh} to the higher-arity case, which measures the validity of a fact as the compatibility between its $n$ values.

{\bf NaLP} \cite{guan2019:nalp} and {\bf RAM} \cite{liu2021:ram} represent an n-ary fact as a set of attribute-value pairs. Then, NaLP employs a convolutional neural network followed by fully connected neural nets to model the relatedness of such attribute-value pairs and accordingly measure the validity of a fact. RAM further encourages to model the relatedness between different attributes and also the relatedness between an attribute and all involved values. 

{\bf HINGE}$\!$ \cite{rosso2020:hinge}  and {\bf NeuInfer}$\!$ \cite{guan2020:NeuInfer} regard an n-ary fact as a primary triple with auxiliary attribute-value pairs. Then they deploy neural modules to measure the validity of the primary triple and its compatibility with each auxiliary description, and combine these modules to obtain the overall score of a fact. As different auxiliary descriptions are modeled independently before aggregation, these two methods show limited ability to model full associations within n-ary facts.
 
{\bf STARE} \cite{galkin2020:stare} is a recently proposed method generalizing graph convolutional networks \cite{kipf2016:gcn} to n-ary relational KGs. It employs a message passing based graph encoder to obtain entity/relation embeddings, and feeds these embeddings to Transformer decoder to score n-ary facts. {\bf Hy-Transformer} \cite{yu2021:hy-transformer} replaces the graph encoder with light-weight embedding processing modules, achieving higher efficiency without sacrificing effectiveness. These two methods employ vanilla Transformer decoders, ignoring specific n-ary structures during decoding.

{\bf n-CP}, {\bf n-TuckER}, and {\bf GETD} \cite{liu2020:getd} are tensor factorization approaches to n-ary link prediction. They all follow RAE and represent each n-ary fact as an $(n\!+\!1)$-tuple. A whole KG can thus be represented as a binary valued $(n+1)$-way tensor $\mathcal{X} \in \{0,1\}^{|\mathcal{R}| \!\times\! |\mathcal{E}| \!\times\! \cdots \!\times\! |\mathcal{E}|}$, where $x=1$ means the corresponding fact is true and $x=0$ otherwise. $\mathcal{X}$ is then decomposed and approximated by a low-rank tensor $\hat{\mathcal{X}}$ that estimates the validity of all facts. Different tensor decomposition strategies can be applied, e.g., n-CP generalizes CP decomposition \cite{kruskal1977:cp} and n-TuckER is built on TuckER \cite{balazevic2019:tucker}. As the tensor representation inherently requires all facts to have the same arity, these methods are not applicable to datasets \mbox{of mixed arities, e.g., JF17K and WikiPeople.}

\subsection{GRAN variants}
We evaluate three variants of GRAN to investigate the impact of modeling graph structure and heterogeneity, including:

{\bf GRAN\scriptsize{-hete}} is the full model introduced above. It uses edge representations defined in Eq.~(\ref{eq:edges}), which encode both graph structure (whether there is an edge) and heterogeneity (which type the edge is).

{\bf GRAN\scriptsize{-homo}} retains graph structure but ignores heterogeneity. There are only two groups of edge attentive biases: $(\mathbf{e}_{ij}^K, \mathbf{e}_{ij}^V) = (\mathbf{0}, \mathbf{0})$ or $(\mathbf{e}_{ij}^K, \mathbf{e}_{ij}^V) = (\mathbf{e}^K, \mathbf{e}^V)$. The former is used if there is no edge between vertices $i$ and $j$, while the latter is employed whenever the two vertices are linked, irrespective of the type of the edge between them. This in essence views an n-ary fact as a homogeneous graph where all edges are of the same type. 

{\bf GRAN\scriptsize{-complete}} considers neither graph structure nor heterogeneity. It simply sets $(\mathbf{e}_{ij}^K, \mathbf{e}_{ij}^V) \!=\! (\mathbf{0}, \mathbf{0})$ for all vertex pairs. The edge-biased attention thus degenerates to the conventional one used in Transformer, which captures only global dependencies between vertices. This in essence regards an n-ary fact as a complete graph in which any two vertices are connected by an (untyped) edge. STARE and Hy-Transformer are most similar to this variant.

We use the following configurations for all variants of GRAN: $L\!=\!12$ graph attention layers, $H\!=$ $4$ attention heads, hidden size $d=256$, batch size $b\!=\!1024$, and learning rate $\eta = 5e\!-\!4$, fixed across all the datasets. Besides, on each dataset, we tune entity/relation label smoothing rate $\epsilon^{(e)}$/$\epsilon^{(r)}$, drop-out rate $\rho$, and training epochs $\tau$ in their respective ranges. The optimal configuration is determined by dev MRR. We leave the tuning ranges and optimal values of these hyperparameters to Appendix~\ref{sec:append-hyperparameter}. After determining the optimal configuration on each dataset, we train with a combination of the train and dev splits and evaluate on the test split, as practiced in \cite{galkin2020:stare}.

\subsection{Evaluation protocol and metrics}
During evaluation, we distinguish between entity prediction and relation prediction. Take entity prediction as an example. For each test n-ary fact, we replace one of its entities (i.e., subject, object, or an auxiliary value) with the special token [MASK], feed the masked graph into GRAN, and obtain a predicted distribution of the answer over all entities $\in \mathcal{E}$. Then we sort the distribution probabilities in descending order and get the rank of the correct answer. During ranking, we ignore facts that already exist in the train, dev, or test split. We repeat this whole procedure for all specified entities in the test fact, and report MRR and Hits@$k$ for $k=1,10$ aggregated on the test split. MRR is the average of reciprocal rankings, and Hits@$k$ is the proportion of top $k$ rankings (abbreviated as H@$k$). The same evaluation protocol and metrics also apply to relation prediction, where a relation can be either the primary relation or an auxiliary attribute.

\begin{table*}[t]
\small\centering\setlength{\tabcolsep}{1pt}
\begin{tabular*}{1 \textwidth}{@{\extracolsep{\fill}}@{}lccclccclccclccc@{}}
\toprule
\multirow{3}*{} & \multicolumn{3}{c}{JF17K} && \multicolumn{3}{c}{JF17K} && \multicolumn{3}{c}{WikiPeople} && \multicolumn{3}{c}{WikiPeople$^-$} \\
& \multicolumn{3}{c}{All Entities} && \multicolumn{3}{c}{Subject/Object} && \multicolumn{3}{c}{All Entities} && \multicolumn{3}{c}{Subject/Object} \\
\cmidrule{2-4} \cmidrule{6-8} \cmidrule{10-12} \cmidrule{14-16}
& MRR & H@1 & H@10 && MRR & H@1 & H@10 && MRR & H@1 & H@10 && MRR & H@1 & H@10 \\
\midrule
RAE        & .310 & .219 & .504 && .215 & .215 & .467 && .172 & .102 & .320 && .059 & .059 & .306 \\ 
NaLP       & .366 & .290 &.516  && .221 & .165 & .331 && .338 & .272 & .466 && .408 & .331 & .546 \\
HINGE    & --      & --     & --      && .449 & .361 & .624 && --     & --      & --      && .476 & .415 & .585 \\
NeuInfer  & .517 & .436 & .675 && --     & --      & --      && .350 & .282 & .467 && --     & --     & --      \\
RAM        & .539 & .463 & .690 && --     & --      & --      && .380 & .279 & .539 && --     & --     & --      \\
STARE    & --      & --     & --      && .574 & .496 & .725 && --     & --      & --      && .491 & .398 & {\bf .648} \\
Hy-Transformer & -- & -- & --     && .582 & .501 & .742 && --     & --      & --      && .501 & .426 & .634 \\
\midrule
GRAN\scriptsize{-hete}   & {\bf .656} & {\bf .582} & {\bf .799}  
                                      && {\bf .617} & {\bf .539} & {\bf .770} 
                                      && {\bf .479} & {\bf .410} & {\bf .604} 
                                      && {\bf .503} & {\bf .438} & .620 \\
GRAN\scriptsize{-homo}       & .650 & .576 & .795 && .611 & .533 & .767 && .465 & .386 & .602 && .487 & .410 & .618 \\
GRAN\scriptsize{-complete} & .622 & .546 & .774 && .591 & .510 & .753 && .460 & .381 & .601 && .489 & .413 & .617 \\
\bottomrule
\end{tabular*}
\caption{\label{tab:mixed-arity-entity} Entity prediction results on JF17K and WikiPeople. RAE and NaLP results for predicting all entities are collected from \cite{guan2020:NeuInfer}, and those for predicting the primary subject/object are collected from \cite{rosso2020:hinge}. Other baseline results are collected from their original literatures. Best scores are highlighted in bold, and ``--'' denotes missing scores.}
\end{table*}

\begin{table*}[t]
\small\centering\setlength{\tabcolsep}{1pt}
\begin{tabular*}{1 \textwidth}{@{\extracolsep{\fill}}@{}lccclccclccclccc@{}}
\toprule
\multirow{3}*{} & \multicolumn{3}{c}{JF17K} && \multicolumn{3}{c}{JF17K} && \multicolumn{3}{c}{WikiPeople} && \multicolumn{3}{c}{WikiPeople$^-$} \\
& \multicolumn{3}{c}{All Relations} && \multicolumn{3}{c}{Primary Relation} && \multicolumn{3}{c}{All Relations} && \multicolumn{3}{c}{Primary Relation} \\
\cmidrule{2-4} \cmidrule{6-8} \cmidrule{10-12} \cmidrule{14-16}
& MRR & H@1 & H@10 && MRR & H@1 & H@10 && MRR & H@1 & H@10 && MRR & H@1 & H@10 \\
\midrule
NaLP       & .825 & .762 & .927 && .639 & .547 & .822 && .735 & .595 & .938 && .482 & .320 & .852 \\
HINGE    & --      & --     & --      && .937 & .901 & .989 && --      & --      & --     && .950 & .916 & {\bf .998} \\
NeuInfer  & .861 & .832 & .904 && --     & --      & --      && .765 & .686 & .897 && --     & --     & --      \\
\midrule
GRAN\scriptsize{-hete}   & {\bf .996} & {\bf .993} & {\bf .999}  
                                      && {\bf .992} & {\bf .988} & {\bf .998} 
                                      && {\bf .960} & {\bf .946} & {\bf .977} 
                                      && {\bf .957} & {\bf .942} & .976 \\
GRAN\scriptsize{-homo}       & .980 & .965 & .998 && .964 & .939 & .997 && .940 & .910 & .975 && .932 & .899 & .971 \\
GRAN\scriptsize{-complete} & .979 & .963 & .998 && .963 & .935 & .997 && .940 & .910 & .976 && .935 & .902 & .974 \\
\bottomrule
\end{tabular*}
\caption{\label{tab:mixed-arity-relation} Relation prediction results on JF17K and WikiPeople. Baseline results for predicting all relations taken from \cite{guan2019:nalp,guan2020:NeuInfer}, and those for predicting the primary relation taken from \cite{rosso2020:hinge}. Best scores are highlighted in bold, and ``--'' denotes missing scores.}
\end{table*}

\subsection{Results on datasets of mixed arities}
Table~\ref{tab:mixed-arity-entity} presents entity prediction results on JF17K and the two versions of WikiPeople, which consist of facts with mixed arities.\footnote{Tensor factorization based approaches which require all facts to have the same arity are not applicable here.} We consider two settings: (1) predicting all entities $s, o, v_1, \cdots, v_m$ in an n-ary fact and (2) predicting only the subject $s$ and object $o$. This enables us to make a direct comparison to previous literatures \cite{guan2020:NeuInfer,rosso2020:hinge,galkin2020:stare}. From the results, we can see that (1) The optimal setting of our approach offers consistent and substantial improvements over all the baselines across all the datasets in almost all metrics, showing its significant effectiveness and superiority in entity prediction within n-ary facts. (2) All the variants, including the less expressive GRAN{\scriptsize-homo} and GRAN{\scriptsize-complete}, perform quite well, greatly surpassing the competitive baselines in almost all cases except for the WikiPeople$^-$ dataset. This verifies the superior effectiveness of modeling n-ary facts as whole graphs so as to capture global dependencies between all relations and entities therein. (3) Among the variants, GRAN{\scriptsize-hete} offers the best performance. This demonstrates the necessity and superiority of further modeling specific graph structures and graph heterogeneity, so as to capture local dependencies reflected by typed edges linking relations and entities.

Table~\ref{tab:mixed-arity-relation} further shows relation prediction results on these datasets. Again, to make direct comparison with previous literatures, we consider two settings: (1) predicting all relations including the primary relation $r$ and auxiliary attributes $a_1, \cdots, a_m$ and (2) predicting only the primary relation $r$. Here, on each dataset, GRAN models are fixed to their respective optimal configurations (see Appendix~\ref{sec:append-hyperparameter}) determined in the entity prediction task.\footnote{Relation prediction typically requires much less training epochs than entity prediction according to our initial experiments. But we did not conduct hyperparameter searching for relation prediction, as the configurations we used, though not optimal, perform well enough in this task.} The results show that GRAN variants perform particularly well in relation prediction. Among these variants, GRAN{\scriptsize-hete} performs the best, consistently outperforming the baselines and achieving extremely high performance across all the datasets. This is because relation prediction is, by nature, a relatively easy task due to a small number of candidate answers.

\begin{table*}[t]
\small\centering\setlength{\tabcolsep}{1pt}
\begin{tabular*}{1 \textwidth}{@{\extracolsep{\fill}}@{}lccclccclccclccc@{}}
\toprule
\multirow{3}*{} & \multicolumn{3}{c}{JF17K-3} && \multicolumn{3}{c}{JF17K-4} && \multicolumn{3}{c}{WikiPeople-3} && \multicolumn{3}{c}{WikiPeople-4} \\
& \multicolumn{3}{c}{All Entities} && \multicolumn{3}{c}{All Entities} && \multicolumn{3}{c}{All Entities} && \multicolumn{3}{c}{All Entities} \\
\cmidrule{2-4} \cmidrule{6-8} \cmidrule{10-12} \cmidrule{14-16}
& MRR & H@1 & H@10 && MRR & H@1 & H@10 && MRR & H@1 & H@10 && MRR & H@1 & H@10 \\
\midrule
RAE          & .505 & .430 & .644 && .707 & .636 & .835 && .239 & .168 & .379 && .150 & .080 & .273 \\ 
NaLP        & .515 & .431 &.679  && .719 & .673 & .805 && .301 & .226 & .445 && .342 & .237 & .540 \\
n-CP         & .700 & .635 & .827 && .787 & .733 & .890 && .330 & .250 & .496 && .265 & .169 & .445 \\
n-TuckER & .727 & .664 & .852 && .804 & .748 & .902 && .365 & .274 & .548 && .362 & .246 & .570 \\
GETD       & .732 & .669 & .856 && .810 & .755 & .913 && .373 & .284 & .558 && .386 & .265 & .596 \\
\midrule
GRAN\scriptsize{-hete}        & {\bf .806} & {\bf .759} & {\bf .896} 
                                           && {\bf .852} & {\bf .801} & {\bf .939}
                                           && {\bf .416} & {\bf .325} & {\bf .608} 
                                           && {\bf .431} & {\bf .309} & {\bf .642} \\
GRAN\scriptsize{-homo}      &  .803 & .755 & {\bf .896} && .848 & .795 & .937 && .410 & .315 & .606 && .426 & .305 & .631 \\
GRAN\scriptsize{-complete} & .730 & .670 & .862         && .844 & .794 & .930 && .408 & .314 & .602 && .365 & .248 & .604 \\
\bottomrule
\end{tabular*}
\caption{\label{tab:single-arity-entity} Entity prediction results on the four JF17K and WikiPeople subsets. Baseline results are taken from \cite{liu2020:getd}. Best scores are highlighted in bold.}
\end{table*}

\begin{table*}[t]
\small\centering\setlength{\tabcolsep}{1pt}
\begin{tabular*}{1 \textwidth}{@{\extracolsep{\fill}}@{}lclclcllclclcllclclc@{}}
\toprule
\multirow{3}*{} & \multicolumn{5}{c}{JF17K} &&& \multicolumn{5}{c}{WikiPeople} &&& \multicolumn{5}{c}{WikiPeople$^-$} \\
\cmidrule{2-6} \cmidrule{9-13} \cmidrule{16-20}
& \multicolumn{3}{c}{Subject/Object} && \multicolumn{1}{c}{Values} &&& \multicolumn{3}{c}{Subject/Object} && \multicolumn{1}{c}{Values} &&& \multicolumn{3}{c}{Subject/Object} && \multicolumn{1}{c}{Values} \\
\cmidrule{2-4} \cmidrule{6-6} \cmidrule{9-11} \cmidrule{13-13} \cmidrule{16-18} \cmidrule{20-20}
& \multicolumn{1}{c}{$n=2$} && \multicolumn{1}{c}{$n>2$} && \multicolumn{1}{c}{$n>2$} &&& \multicolumn{1}{c}{$n=2$} && \multicolumn{1}{c}{$n>2$} && \multicolumn{1}{c}{$n>2$} &&& \multicolumn{1}{c}{$n=2$} && \multicolumn{1}{c}{$n>2$} && \multicolumn{1}{c}{$n>2$} \\
\midrule
GRAN\scriptsize{-hete}        & .413 && .768 && .758 &&& .495 && .361 && .471 &&& .503 && .505 && .713 \\
GRAN\scriptsize{-homo}      & .409 && .759 && .753 &&& .479 && .354 && .467 &&& .487 && .486 && .690 \\
GRAN\scriptsize{-complete} & .409 && .725 && .705 &&& .478 && .353 && .415 &&& .489 && .480 && .665 \\
\bottomrule
\end{tabular*}
\caption{\label{tab:breakdown} Breakdown performance of the GRAN variants in entity prediction task on JF17K and WikiPeople. Only MRR scores are reported.}
\end{table*}

\subsection{Results on datasets of a single arity}
Table~\ref{tab:single-arity-entity} presents entity prediction results on the four subsets of JF17K and WikiPeople, which consist solely of 3-ary or 4-ary facts. Here, an entity means either the subject, the object, or an attribute value. On these four single-arity subsets, tensor factorization based approaches like n-CP, n-TuckER, and GETD apply quite well and have reported promising performance \cite{liu2020:getd}. From the results, we can observe similar phenomena as from Table~\ref{tab:mixed-arity-entity}. The GRAN variants perform particularly well, all surpassing or at least performing on par with the baselines across the datasets. $\!$And GRAN{\scriptsize -hete}, again, offers the best performance in general among the three variants.

\subsection{Further analysis}
We further look into the breakdown entity prediction performance of the GRAN variants on different arities. More specifically, we group the test split of each dataset into binary ($n\!=\!2$) and n-ary ($n\!>\!2$) categories. Entity prediction means predicting the subject/object for the binary category, or predicting an attribute value in addition for the n-ary category. Table~\ref{tab:breakdown} presents the breakdown MRR scores in all these different cases on JF17K, WikiPeople, and WikiPeople$^-$, with the GRAN variants set to their respective optimal configurations on each dataset (see Appendix~\ref{sec:append-hyperparameter}). Among the variants GRAN{\scriptsize -hete} performs best in all cases, which again verifies the necessity and superiority of modeling n-ary facts as heterogeneous graphs. Ignoring the graph heterogeneity (GRAN{\scriptsize -homo}) or further graph structures (GRAN{\scriptsize -complete}) always leads to worse performance, particularly when predicting auxiliary attribute values in higher-arity facts.

\section{Related work}\label{sec:relatedwork}
\paragraph{Link prediction on binary relational data}
$\!\!$Most previous work of learning with knowledge graphs (KGs) focused on binary relations. Among different binary relational learning techniques, embedding based models have received increasing attention in recent years due to their effectiveness and simplicity. The idea there is to represent symbolic entities and relations in a continuous vector space and measure the validity of a fact in that space. This kind of models can be roughly grouped into three categories: translation distance based \cite{bordes2013:transe,wang2014:transh,sun2019:rotate}, semantic matching based \cite{trouillon2016:complex,balazevic2019:tucker}, and neural network based \cite{dettmers2018:conve,schlichtkrull2017:rgcn}, according to the design of validity scoring functions. We refer readers to \cite{nickel2016:review,wang2017:review,ji2021:review} for thorough reviews of the literature. 

\paragraph{Link prediction on n-ary relational data}
Since binary relations oversimplify the complex nature of the data stored in KGs, a few recent studies have started to explore learning and reasoning with n-ary relational data ($n>2$), in particular via embedding based approaches. Most of these studies represent n-ary facts as tuples of pre-defined relations with corresponding attribute values, and generalize binary relational learning methods to the n-ary case, e.g., m-TransH \cite{wen2016:m-transh} and RAE \cite{zhang2018:rae} generalize TransH \cite{wang2014:transh}, a translation distance based embedding model for binary relations, while n-CP, n-TuckER, and GETD \cite{liu2020:getd} generalize 3-way tensor decomposition techniques to the higher-arity case. NaLP \cite{guan2019:nalp} and RAM \cite{liu2021:ram} are slightly different approaches which represent n-ary facts directly as groups of attribute-value pairs and then model relatedness between such attributes and values. In these approaches, however, attributes of an n-ary fact are assumed to be equally important, which is often not the case in reality. \citeauthor{rosso2020:hinge} \shortcite{rosso2020:hinge} and \citeauthor{guan2020:NeuInfer} \shortcite{guan2020:NeuInfer} therefore proposed to represent n-ary facts as primary triples coupled with auxiliary attribute-value pairs, which naturally discriminates the importance of different attributes. The overall validity of a fact is then measured by the validity of the primary triple and its compatibility with each attribute-value pair. STARE \cite{galkin2020:stare} follows the same representation form of n-ary facts, and generalizes graph convolutional networks to n-ary relational KGs to learn entity and relation embeddings. These embeddings are then fed into a Transformer decoder to score n-ary facts. Nevertheless, during the decoding process STARE takes into account solely global dependencies and ignores the specific n-ary structure of a given fact.

\paragraph{Transformer and its extensions} Transformer \cite{vaswani2017:transformer} was initially devised as an encoder-decoder architecture for machine translation, and quickly received broad attention across all areas of natural language processing \cite{radford2018:gpt,devlin2019:bert,yang2019:xlnet}. Transformer uses neither convolution nor recurrence, but instead is built entirely with (self-) attention layers. Recently, there has been a lot of interest in modifying this attention to further meet various desired specifications, e.g., to encode syntax trees \cite{strubell2018:lisa,wang2019:treetransformer}, character-word lattice structures \cite{li2020:flat}, as well as relative positions between words \cite{shaw2018:relativeposition,wang2019:relativeposition}. There are also a few recent attempts that apply vanilla Transformer \cite{wang2019:coke} or hierarchical Transformer \cite{chen2020:hitter} to KGs, but mainly restricted to binary relations and deployed with conventional attention. This work, in contrast, deals with higher-arity relational data represented as heterogeneous graphs, and employs modified attention to encode graph structure and heterogeneity.

\section{Conclusion}\label{sec:conclusion}
This paper studies link prediction on higher-arity relational facts and presents a graph-based approach to this task. For each given n-ary fact, our approach (1) represents the fact as a heterogeneous graph in which the semantics of the fact are fully retained; (2) models the graph using fully-connected attention with edge-aware attentive biases so as to capture both local and global dependencies within the given fact. By modeling an n-ary fact as a whole graph, our approach can more effectively capture entity relation associations therein, which is crucial for inference on such facts. Link prediction results on a variety of n-ary relational benchmarks demonstrate the significant effectiveness and superiority of our approach. 

As future work, we would like to (1) verify the effectiveness of GRAN on newly introduced benchmarks such as WD50K \cite{galkin2020:stare} and FB-AUTO \cite{fatemi2020:hype}; (2) investigate the usefulness of specific modules, e.g., positional embeddings and various forms of attentive biases in GRAN; and (3) integrate other types of data in a KG, e.g., entities's textual descriptions, for better n-ary link prediction. 

\section*{Acknowledgments}
We would like to thank the anonymous reviewers for their insightful suggestions. This work is supported by the National Key Research and Development Program of China (No.2020AAA0109400) and the Natural Science Foundation of China (No. 61876223).

\bibliography{acl2021}

\begin{thebibliography}{42}
\expandafter\ifx\csname natexlab\endcsname\relax\def\natexlab#1{#1}\fi

\bibitem[{Ba et~al.(2016)Ba, Kiros, and Hinton}]{ba2016:layernorm}
Jimmy~Lei Ba, Jamie~Ryan Kiros, and Geoffrey~E Hinton. 2016.
\newblock Layer normalization.
\newblock \emph{arXiv preprint arXiv:1607.06450}.

\bibitem[{Balazevic et~al.(2019)Balazevic, Allen, and
  Hospedales}]{balazevic2019:tucker}
Ivana Balazevic, Carl Allen, and Timothy Hospedales. 2019.
\newblock {TuckER}: Tensor factorization for knowledge graph completion.
\newblock In \emph{Proceedings of the 2019 Conference on Empirical Methods in
  Natural Language Processing and the 9th International Joint Conference on
  Natural Language Processing}, pages 5188--5197.

\bibitem[{Bollacker et~al.(2008)Bollacker, Evans, Paritosh, Sturge, and
  Taylor}]{bollacker2008:Freebase}
Kurt Bollacker, Colin Evans, Praveen Paritosh, Tim Sturge, and Jamie Taylor.
  2008.
\newblock Freebase: A collaboratively created graph database for structuring
  human knowledge.
\newblock In \emph{Proceedings of the 2008 ACM SIGMOD International Conference
  on Management of Data}, pages 1247--1250.

\bibitem[{Bordes et~al.(2013)Bordes, Usunier, Garcia-Duran, Weston, and
  Yakhnenko}]{bordes2013:transe}
Antoine Bordes, Nicolas Usunier, Alberto Garcia-Duran, Jason Weston, and Oksana
  Yakhnenko. 2013.
\newblock Translating embeddings for modeling multi-relational data.
\newblock In \emph{Advances in Neural Information Processing Systems}, pages
  2787--2795.

\bibitem[{Chen et~al.(2020)Chen, Liu, Gao, Jiao, Zhang, and
  Ji}]{chen2020:hitter}
Sanxing Chen, Xiaodong Liu, Jianfeng Gao, Jian Jiao, Ruofei Zhang, and Yangfeng
  Ji. 2020.
\newblock {HittER}: Hierarchical {T}ransformers for knowledge graph embeddings.
\newblock \emph{arXiv preprint arXiv:2008.12813}.

\bibitem[{Dettmers et~al.(2018)Dettmers, Minervini, Stenetorp, and
  Riedel}]{dettmers2018:conve}
Tim Dettmers, Pasquale Minervini, Pontus Stenetorp, and Sebastian Riedel. 2018.
\newblock Convolutional {2D} knowledge graph embeddings.
\newblock In \emph{Proceedings of the Thirty-Second AAAI Conference on
  Artificial Intelligence}, pages 1811--1818.

\bibitem[{Devlin et~al.(2019)Devlin, Chang, Lee, and
  Toutanova}]{devlin2019:bert}
Jacob Devlin, Ming-Wei Chang, Kenton Lee, and Kristina Toutanova. 2019.
\newblock {BERT}: Pre-training of deep bidirectional transformers for language
  understanding.
\newblock In \emph{Proceedings of the 2019 Conference of the North American
  Chapter of the Association for Computational Linguistics: Human Language
  Technologies}, pages 4171--4186.

\bibitem[{Dong et~al.(2014)Dong, Gabrilovich, Heitz, Horn, Lao, Murphy,
  Strohmann, Sun, and Zhang}]{dong2014:KnowledgeVault}
Xin Dong, Evgeniy Gabrilovich, Geremy Heitz, Wilko Horn, Ni~Lao, Kevin Murphy,
  Thomas Strohmann, Shaohua Sun, and Wei Zhang. 2014.
\newblock Knowledge vault: A web-scale approach to probabilistic knowledge
  fusion.
\newblock In \emph{Proceedings of the 20th ACM SIGKDD International Conference
  on Knowledge Discovery and Data Mining}, pages 601--610.

\bibitem[{Fatemi et~al.(2020)Fatemi, Taslakian, Vazquez, and
  Poole}]{fatemi2020:hype}
Bahare Fatemi, Perouz Taslakian, David Vazquez, and David Poole. 2020.
\newblock Knowledge hypergraphs: Prediction beyond binary relations.
\newblock In \emph{Proceedings of the Twenty-Ninth International Joint
  Conference on Artificial Intelligence}, pages 2191--2197.

\bibitem[{Galkin et~al.(2020)Galkin, Trivedi, Maheshwari, Usbeck, and
  Lehmann}]{galkin2020:stare}
Mikhail Galkin, Priyansh Trivedi, Gaurav Maheshwari, Ricardo Usbeck, and Jens
  Lehmann. 2020.
\newblock Message passing for hyper-relational knowledge graphs.
\newblock In \emph{Proceedings of the 2020 Conference on Empirical Methods in
  Natural Language Processing}, pages 7346--7359.

\bibitem[{Guan et~al.(2020)Guan, Jin, Guo, Wang, and Cheng}]{guan2020:NeuInfer}
Saiping Guan, Xiaolong Jin, Jiafeng Guo, Yuanzhuo Wang, and Xueqi Cheng. 2020.
\newblock {NeuInfer: Knowledge inference on n-ary facts}.
\newblock In \emph{Proceedings of the 58th Annual Meeting of the Association
  for Computational Linguistics}, pages 6141--6151.

\bibitem[{Guan et~al.(2019)Guan, Jin, Wang, and Cheng}]{guan2019:nalp}
Saiping Guan, Xiaolong Jin, Yuanzhuo Wang, and Xueqi Cheng. 2019.
\newblock Link prediction on n-ary relational data.
\newblock In \emph{Proceedings of the 2019 World Wide Web Conference}, pages
  583--593.

\bibitem[{He et~al.(2016)He, Zhang, Ren, and Sun}]{he2016:resnet}
Kaiming He, Xiangyu Zhang, Shaoqing Ren, and Jian Sun. 2016.
\newblock Deep residual learning for image recognition.
\newblock In \emph{Proceedings of the IEEE Conference on Computer Vision and
  Pattern Recognition}, pages 770--778.

\bibitem[{Hendrycks and Gimpel(2016)}]{hendrycks2016:gelu}
Dan Hendrycks and Kevin Gimpel. 2016.
\newblock Gaussian error linear units ({GELU}s).
\newblock \emph{arXiv preprint arXiv:1606.08415}.

\bibitem[{Ji et~al.(2021)Ji, Pan, Cambria, Marttinen, and
  Philip}]{ji2021:review}
Shaoxiong Ji, Shirui Pan, Erik Cambria, Pekka Marttinen, and S~Yu Philip. 2021.
\newblock A survey on knowledge graphs: Representation, acquisition, and
  applications.
\newblock \emph{IEEE Transactions on Neural Networks and Learning Systems}.

\bibitem[{Kingma and Ba(2015)}]{kingma2014:adam}
Diederik~P Kingma and Jimmy Ba. 2015.
\newblock Adam: A method for stochastic optimization.
\newblock In \emph{International Conference for Learning Representations}.

\bibitem[{Kipf and Welling(2017)}]{kipf2016:gcn}
Thomas~N Kipf and Max Welling. 2017.
\newblock Semi-supervised classification with graph convolutional networks.
\newblock In \emph{International Conference for Learning Representations}.

\bibitem[{Kruskal(1977)}]{kruskal1977:cp}
Joseph~B Kruskal. 1977.
\newblock Three-way arrays: Rank and uniqueness of trilinear decompositions,
  with application to arithmetic complexity and statistics.
\newblock \emph{Linear Algebra and Its Applications}, 18(2):95--138.

\bibitem[{Li et~al.(2020)Li, Yan, Qiu, and Huang}]{li2020:flat}
Xiaonan Li, Hang Yan, Xipeng Qiu, and Xuanjing Huang. 2020.
\newblock {FLAT}: {C}hinese {NER} using flat-lattice {T}ransformer.
\newblock In \emph{Proceedings of the 58th Annual Meeting of the Association
  for Computational Linguistics}, pages 6836--6842.

\bibitem[{Liu et~al.(2020)Liu, Yao, and Li}]{liu2020:getd}
Yu~Liu, Quanming Yao, and Yong Li. 2020.
\newblock Generalizing tensor decomposition for n-ary relational knowledge
  bases.
\newblock In \emph{Proceedings of The Web Conference 2020}, pages 1104--1114.

\bibitem[{Liu et~al.(2021)Liu, Yao, and Li}]{liu2021:ram}
Yu~Liu, Quanming Yao, and Yong Li. 2021.
\newblock Role-aware modeling for n-ary relational knowledge bases.
\newblock \emph{arXiv preprint arXiv:2104.09780}.

\bibitem[{Nickel et~al.(2016)Nickel, Murphy, Tresp, and
  Gabrilovich}]{nickel2016:review}
Maximilian Nickel, Kevin Murphy, Volker Tresp, and Evgeniy Gabrilovich. 2016.
\newblock A review of relational machine learning for knowledge graphs.
\newblock \emph{Proceedings of the IEEE}, 104(1):11--33.

\bibitem[{Radford et~al.(2018)Radford, Narasimhan, Salimans, and
  Sutskever}]{radford2018:gpt}
Alec Radford, Karthik Narasimhan, Time Salimans, and Ilya Sutskever. 2018.
\newblock Improving language understanding by generative pre-training.
\newblock Technical report, OpenAI.

\bibitem[{Rosso et~al.(2020)Rosso, Yang, and
  Cudr{\'e}-Mauroux}]{rosso2020:hinge}
Paolo Rosso, Dingqi Yang, and Philippe Cudr{\'e}-Mauroux. 2020.
\newblock Beyond triplets: Hyper-relational knowledge graph embedding for link
  prediction.
\newblock In \emph{Proceedings of The Web Conference 2020}, pages 1885--1896.

\bibitem[{Schlichtkrull et~al.(2018)Schlichtkrull, Kipf, Bloem, Van Den~Berg,
  Titov, and Welling}]{schlichtkrull2017:rgcn}
Michael Schlichtkrull, Thomas~N Kipf, Peter Bloem, Rianne Van Den~Berg, Ivan
  Titov, and Max Welling. 2018.
\newblock Modeling relational data with graph convolutional networks.
\newblock In \emph{European Semantic Web Conference}, pages 593--607.

\bibitem[{Shaw et~al.(2018)Shaw, Uszkoreit, and
  Vaswani}]{shaw2018:relativeposition}
Peter Shaw, Jakob Uszkoreit, and Ashish Vaswani. 2018.
\newblock Self-attention with relative position representations.
\newblock In \emph{Proceedings of the 2018 Conference of the North American
  Chapter of the Association for Computational Linguistics: Human Language
  Technologies}, pages 464--468.

\bibitem[{Strubell et~al.(2018)Strubell, Verga, Andor, Weiss, and
  McCallum}]{strubell2018:lisa}
Emma Strubell, Patrick Verga, Daniel Andor, David Weiss, and Andrew McCallum.
  2018.
\newblock Linguistically-informed self-attention for semantic role labeling.
\newblock In \emph{Proceedings of the 2018 Conference on Empirical Methods in
  Natural Language Processing}, pages 5027--5038.

\bibitem[{Sun et~al.(2019)Sun, Deng, Nie, and Tang}]{sun2019:rotate}
Zhiqing Sun, Zhi-Hong Deng, Jian-Yun Nie, and Jian Tang. 2019.
\newblock {RotatE}: Knowledge graph embedding by relational rotation in complex
  space.
\newblock In \emph{International Conference for Learning Representations}.

\bibitem[{Trouillon et~al.(2016)Trouillon, Welbl, Riedel, Gaussier, and
  Bouchard}]{trouillon2016:complex}
Th{\'e}o Trouillon, Johannes Welbl, Sebastian Riedel, {\'E}ric Gaussier, and
  Guillaume Bouchard. 2016.
\newblock Complex embeddings for simple link prediction.
\newblock In \emph{Proceedings of the 33rd International Conference on Machine
  Learning}, pages 2071--2080.

\bibitem[{Vaswani et~al.(2017)Vaswani, Shazeer, Parmar, Uszkoreit, Jones,
  Gomez, Kaiser, and Polosukhin}]{vaswani2017:transformer}
Ashish Vaswani, Noam Shazeer, Niki Parmar, Jakob Uszkoreit, Llion Jones,
  Aidan~N Gomez, {\L}ukasz Kaiser, and Illia Polosukhin. 2017.
\newblock Attention is all you need.
\newblock In \emph{Advances in Neural Information Processing Systems}, pages
  5998--6008.

\bibitem[{Vrande{\v{c}}i{\'c} and Kr{\"o}tzsch(2014)}]{vrandevcic2014:Wikidata}
Denny Vrande{\v{c}}i{\'c} and Markus Kr{\"o}tzsch. 2014.
\newblock Wikidata: A free collaborative knowledge base.
\newblock \emph{Communications of the ACM}, 57(10):78--85.

\bibitem[{Wang et~al.(2019{\natexlab{a}})Wang, Tan, Yu, Chang, Wang, Xu, Guo,
  and Potdar}]{wang2019:relativeposition}
Haoyu Wang, Ming Tan, Mo~Yu, Shiyu Chang, Dakuo Wang, Kun Xu, Xiaoxiao Guo, and
  Saloni Potdar. 2019{\natexlab{a}}.
\newblock Extracting multiple-relations in one-pass with pre-trained
  {Transformers}.
\newblock In \emph{Proceedings of the 57th Annual Meeting of the Association
  for Computational Linguistics}, pages 1371--1377.

\bibitem[{Wang et~al.(2019{\natexlab{b}})Wang, Huang, Wang, Dai, Jiang, Liu,
  Lyu, Zhu, and Wu}]{wang2019:coke}
Quan Wang, Pingping Huang, Haifeng Wang, Songtai Dai, Wenbin Jiang, Jing Liu,
  Yajuan Lyu, Yong Zhu, and Hua Wu. 2019{\natexlab{b}}.
\newblock {CoKE}: Contextualized knowledge graph embedding.
\newblock \emph{arXiv preprint arXiv:1911.02168}.

\bibitem[{Wang et~al.(2017)Wang, Mao, Wang, and Guo}]{wang2017:review}
Quan Wang, Zhendong Mao, Bin Wang, and Li~Guo. 2017.
\newblock Knowledge graph embedding: A survey of approaches and applications.
\newblock \emph{IEEE Transactions on Knowledge and Data Engineering},
  29(12):2724--2743.

\bibitem[{Wang et~al.(2019{\natexlab{c}})Wang, Lee, and
  Chen}]{wang2019:treetransformer}
Yaushian Wang, Hung-Yi Lee, and Yun-Nung Chen. 2019{\natexlab{c}}.
\newblock Tree {T}ransformer: Integrating tree structures into self-attention.
\newblock In \emph{Proceedings of the 2019 Conference on Empirical Methods in
  Natural Language Processing and the 9th International Joint Conference on
  Natural Language Processing}, pages 1061--1070.

\bibitem[{Wang et~al.(2014)Wang, Zhang, Feng, and Chen}]{wang2014:transh}
Zhen Wang, Jianwen Zhang, Jianlin Feng, and Zheng Chen. 2014.
\newblock Knowledge graph embedding by translating on hyperplanes.
\newblock In \emph{Proceedings of the Twenty-Eighth AAAI Conference on
  Artificial Intelligence}, pages 1112--1119.

\bibitem[{Wen et~al.(2016)Wen, Li, Mao, Chen, and Zhang}]{wen2016:m-transh}
Jianfeng Wen, Jianxin Li, Yongyi Mao, Shini Chen, and Richong Zhang. 2016.
\newblock On the representation and embedding of knowledge bases beyond binary
  relations.
\newblock In \emph{Proceedings of the Twenty-Fifth International Joint
  Conference on Artificial Intelligence}, pages 1300--1307.

\bibitem[{West et~al.(2014)West, Gabrilovich, Murphy, Sun, Gupta, and
  Lin}]{west2014:Incomplete}
Robert West, Evgeniy Gabrilovich, Kevin Murphy, Shaohua Sun, Rahul Gupta, and
  Dekang Lin. 2014.
\newblock Knowledge base completion via search-based question answering.
\newblock In \emph{Proceedings of the 23rd International Conference on World
  Wide Web}, pages 515--526.

\bibitem[{Xu et~al.(2021)Xu, Wang, Lyu, Zhu, and Mao}]{xu2021:ssan}
Benfeng Xu, Quan Wang, Yajuan Lyu, Yong Zhu, and Zhendong Mao. 2021.
\newblock Entity structure within and throughout: Modeling mention dependencies
  for document-level relation extraction.
\newblock \emph{arXiv preprint arXiv:2102.10249}.

\bibitem[{Yang et~al.(2019)Yang, Dai, Yang, Carbonell, Salakhutdinov, and
  Le}]{yang2019:xlnet}
Zhilin Yang, Zihang Dai, Yiming Yang, Jaime Carbonell, Russ~R Salakhutdinov,
  and Quoc~V Le. 2019.
\newblock {XLNet}: Generalized autoregressive pretraining for language
  understanding.
\newblock In \emph{Advances in Neural Information Processing Systems}, pages
  5753--5763.

\bibitem[{Yu and Yang(2021)}]{yu2021:hy-transformer}
Donghan Yu and Yiming Yang. 2021.
\newblock Improving hyper-relational knowledge graph completion.
\newblock \emph{arXiv preprint arXiv:2104.08167}.

\bibitem[{Zhang et~al.(2018)Zhang, Li, Mei, and Mao}]{zhang2018:rae}
Richong Zhang, Junpeng Li, Jiajie Mei, and Yongyi Mao. 2018.
\newblock Scalable instance reconstruction in knowledge bases via relatedness
  affiliated embedding.
\newblock In \emph{Proceedings of the 2018 World Wide Web Conference}, pages
  1185--1194.

\end{thebibliography}
\bibliographystyle{acl_natbib}

\appendix

\section{Graph attention layers}\label{sec:append-gat}
After the input embedding layer, we employ a stack of $L$ identical graph attention layers to encode the input graph before making final predictions. These graph attention layers generally follow the design of Transformer encoder \cite{vaswani2017:transformer}, each of which consists of two sub-layers, i.e., an edge-biased fully-connected attention sub-layer followed by an element-wise feed-forward sub-layer. The attention sub-layer, as illustrated in Section 3.2, relates different vertices of the input graph to update its vertex representations. It computes attention in an edge-biased fully-connected fashion, which thus is able to capture both global and local dependencies within the graph. The feed-forward sub-layer is composed of two linear transformations with a GELU activation~\cite{hendrycks2016:gelu} in between, applied to each element/vertex separately and identically. We further introduce residual connections~\cite{he2016:resnet} and layer normalization \cite{ba2016:layernorm} around each graph attention layer and its sub-layers. To facilitate these residual connections, all the layers and their sub-layers produce outputs of the same dimension $d$.

\section{Hyperparameter settings}\label{sec:append-hyperparameter}
We use the following hyperparameter settings for GRAN: $L\!=\!12$ graph attention layers, $H\!=\!4$ attention heads, hidden size $d\!=\!256$, batch size $b\!=\!\!1024$, and learning rate $\eta \!=\! 5e\!-\!4$. These configurations are fixed across all the datasets. Besides, on each dataset we tune the following hyperparameters in their respective ranges:
\begin{itemize}
\item entity label smoothing rate $\epsilon^{(e)} \in \{.0, .1, .2,$ $.3, .4, .5, .6, .7, .8, .9\}$;
\item relation label smoothing rate $\epsilon^{(r)} \in \{.0, .1, .2,$ $.3, .4, .5\}$;
\item dropout rate $\rho \in \{.1, .2, .3, .4, .5\}$;
\item training epochs $\tau$ from 20 to 200 in steps of 20 on all the datasets.
\end{itemize}
We determine the optimal configuration for GRAN{\scriptsize -hete} by dev MRR of entity prediction on each dataset. And then we directly set GRAN{\scriptsize -homo} and GRAN{\scriptsize -complete} to the same configuration. Table~\ref{tab:hyperparameters} presents the optimal configuration on each dataset.

\begin{table}[h]
\small\centering\setlength{\tabcolsep}{1pt}
\begin{tabular*}{0.47 \textwidth}{@{\extracolsep{\fill}}@{}lcccccc@{}}
\toprule
& $\epsilon^{(e)}$ & $\epsilon^{(r)}$ & $\rho$ & $\tau$ \\
\midrule
JF17K                & .9 & .0 & .2 & 160 \\
WikiPeople        & .2 & .2 & .1 & 200 \\
WikiPeople$^-$ & .2 & .1 & .1 & 160 \\
\midrule
JF17K-3        & .8 & .2 & .2 & 180 \\
JF17K-4        & .8 & .0 & .3 & 160 \\
WikiPeople-3 & .8 & .4 & .3 & 100 \\
WikiPeople-4 & .8 & .4 & .3 & 100 \\
\bottomrule
\end{tabular*}
\caption{\label{tab:hyperparameters} Optimal configuration of the GRAN variants on each dataset.}
\end{table}

\section{Infrastructure and runtime}\label{sec:append-runtime}
We train all the GRAN variants on one 16G V100 GPU. With the hyperparameter settings specified in Appendix~\ref{sec:append-hyperparameter}, it takes about 3 hours to finish training and evaluation on JF17K, 17 hours on Wikipeople, 10 hours on Wikipeople$^-$, 1 hour on JF17K-3, 0.5 hour on JF17K-4, Wikipeople-3, and WikiPeople-4. This runtime covers the whole training and evaluation process. Compared to previous methods like HINGE \cite{rosso2020:hinge} and NeuInfer \cite{guan2020:NeuInfer} which score individual facts and learn from positive-negative pairs, GRAN directly scores each target answer against all candidates in a single pass and drastically speeds up evaluation. GRAN is also much more efficient than STARE \cite{galkin2020:stare}, which is a graph encoder plus Transformer decoder architecture. By eliminating the computational heavy graph encoder, GRAN requires significantly less running time but still achieves better performance than STARE, e.g., GRAN{\scriptsize -hete} achieves .617 MRR within 3 hours while STARE takes about 10 hours to achieve .574 MRR on JF17K; GRAN{\scriptsize -hete} achieves .503 MRR within 10 hours but STARE takes about 4 days to achieve a similar MRR on Wikipeople$^-$ (which is 9-10 times slower).

\end{document}